\documentclass[10pt,twocolumn,letterpaper]{article}

\usepackage{iccv}
\usepackage{times}
\usepackage{epsfig}
\usepackage{graphicx}
\usepackage{amsmath}
\usepackage{amssymb}
\usepackage[american]{babel}
\usepackage[ruled,linesnumbered]{algorithm2e}
\usepackage{booktabs}
\usepackage{float}
\usepackage{multirow}
\usepackage{indentfirst} 
\usepackage{amssymb}
\usepackage[table]{xcolor}
\usepackage{colortbl}
\usepackage{subfigure}
\usepackage{appendix}
\usepackage{indentfirst} 
\usepackage{chngcntr}
\usepackage[accsupp]{axessibility}

\usepackage[breaklinks=true,bookmarks=false]{hyperref}

\usepackage[capitalize]{cleveref}
\crefname{section}{Sec.}{Secs.}
\Crefname{section}{Section}{Sections}
\Crefname{table}{Table}{Tables}
\crefname{table}{Table}{Tables}

\iccvfinalcopy 

\def\iccvPaperID{****} 

\ificcvfinal\fi

\begin{document}

\title{When Epipolar Constraint Meets Non-local Operators in Multi-View Stereo
}

\author{
  Tianqi Liu \quad
  Xinyi Ye \quad
  Weiyue Zhao \quad
  Zhiyu Pan \quad
  Min Shi\thanks{Corresponding author} \quad
  Zhiguo Cao \\
  Key Laboratory of Image Processing and Intelligent Control, Ministry of Education; School of Artificial \\
  Intelligence and Automation, Huazhong University of Science and Technology, Wuhan 430074, China \\
  {\tt\small \{tianqiliu,xinyiye,zhaoweiyue,zhiyupan,min\_shi,zgcao\}@hust.edu.cn}
}

\maketitle
\ificcvfinal\thispagestyle{empty}\fi

\begin{abstract}
  Learning-based multi-view stereo (MVS) method heavily relies on feature matching, which requires distinctive and descriptive representations.
  An effective solution is to apply non-local feature aggregation, \textit{e.g.}, Transformer. Albeit useful, these techniques introduce heavy computation overheads for MVS. Each pixel densely attends to the whole image.
  In contrast, we propose to constrain non-local feature augmentation within a pair of lines: each point only attends the corresponding pair of epipolar lines.
  Our idea takes inspiration from the classic epipolar geometry, which shows that one point with different depth hypotheses will be projected to the epipolar line on the other view. 
  This constraint reduces the 2D search space into the epipolar line in stereo matching. Similarly, this suggests that the matching of MVS is to distinguish a series of points lying on the same line.
  Inspired by this point-to-line search, we devise a line-to-point non-local augmentation strategy. We first devise an optimized searching algorithm to split the 2D feature maps into epipolar line pairs.
  Then, an Epipolar Transformer (ET) performs non-local feature augmentation among epipolar line pairs.
  We incorporate the ET into a learning-based MVS baseline, named ET-MVSNet. ET-MVSNet achieves state-of-the-art reconstruction performance on both the DTU and Tanks-and-Temples benchmark with high efficiency. Code is available at \href{https://github.com/TQTQliu/ET-MVSNet}{https://github.com/TQTQliu/ET-MVSNet}.

\end{abstract}

\section{Introduction}

\begin{figure}
    \includegraphics[scale=0.32]{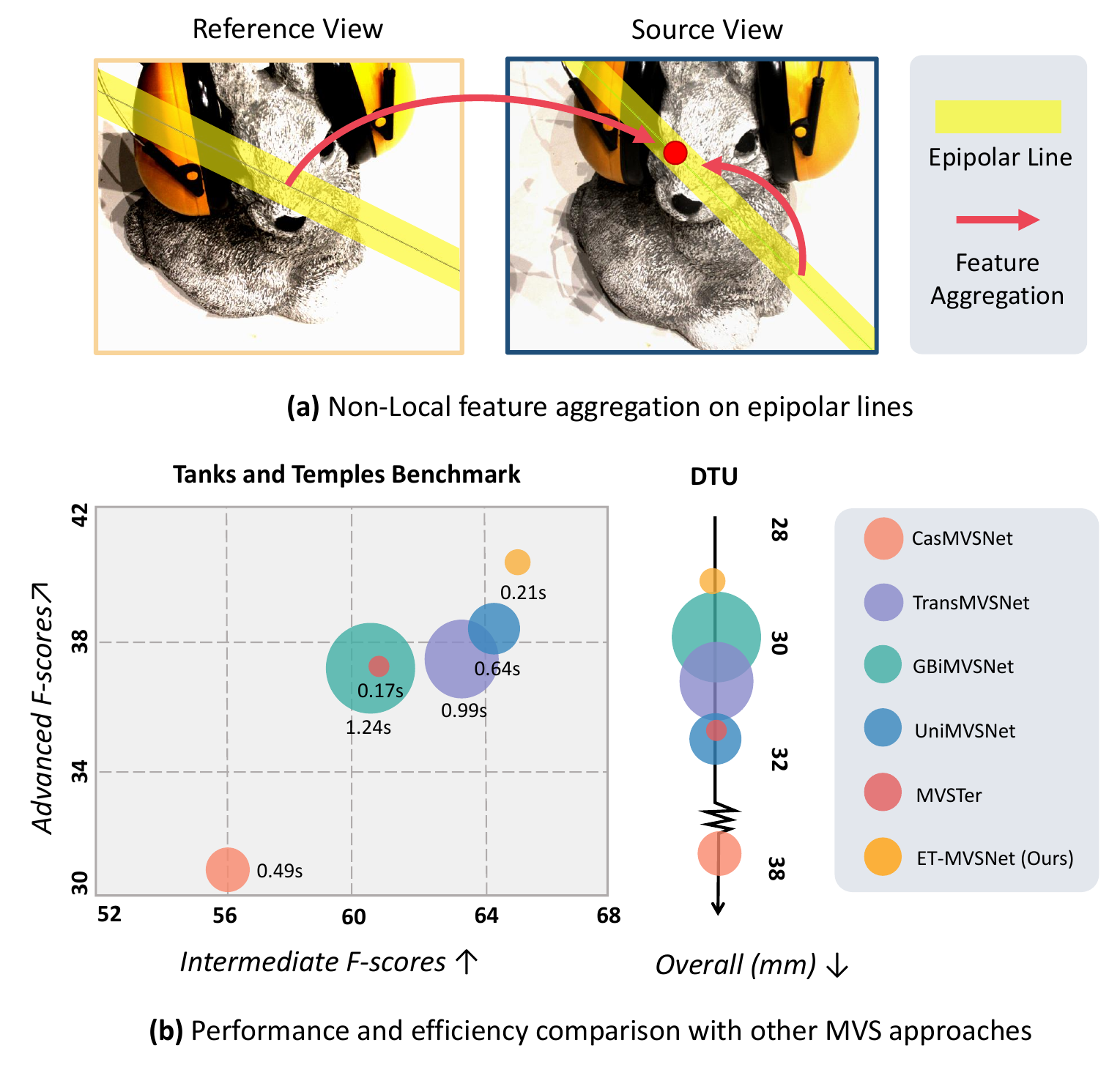}
    \caption{\textbf{Our methods and results.} (a) Inspired by classic epipolar geometry in stereo matching, we propose an efficient non-local feature aggregation paradigm that aggregates features on the corresponding epipolar line pairs. (b) We instantiate the proposed non-local mechanism with a Transformer, termed Epipolar Transformer (ET), which attains state-of-the-art performance with limited computation overhead.}
    \label{fig1}
\end{figure}

As a fundamental topic in computer vision, Multi-view stereo (MVS) aims to reconstruct dense 3D representations from multiple calibrated images given corresponding camera parameters. MVS can facilitate many applications such as automatic driving and augmented reality~\cite{furukawa2015multi}. A critical part of MVS is to match pixels to find the corresponding points. This matching process heavily relies on the feature representation: corresponding points in different views should be close in the embedding space. Traditional MVS approaches~\cite{fua1995object,gipuma,colmap,schonberger2016pixelwise} adopt hand-craft feature representation, which faces challenges on repetitive-pattern, weak-texture, and reflective regions.

To remedy the weak representation of the traditional MVS approaches, learning-based methods are proposed and show superior performance.
Learning-based MVS approaches adopt deep neural networks to encode features, which forms the foundation for the subsequent feature matching and depth generation. 
Powerful as convolutional neural networks (CNNs) are, they still prefer to aggregate local context~\cite{luo2016understanding}, which may be less accurate for matching. 
One feasible solution is to adopt non-local feature augmentation strategy with large receptive field and flexible feature aggregation, which is crucial for robust and descriptive feature representations. 

To this end, techniques such as deformable convolution~\cite{dai2017deformable}, attention~\cite{bahdanau2014neural}, and Transformers~\cite{vaswani2017attention}, are introduced to the feature encoding of MVS.
Albeit useful, these general-purpose non-local operators introduce huge computation overhead for MVS, as each point need to densely attend the reference and source images. 
%
In this case, distractive or irrelevant features can also be attended which are harmful to the feature matching, \textit{e.g.}, repetitive patterns shown in Fig.~\ref{fig6}. 
This raises a fundamental question: \textit{where and how to mine non-local context for MVS?} 
Inspired by the classic epipolar geometry, we propose to constrain the non-local feature aggregation within the epipolar lines, which enjoys both efficiency and descriptive representation. 

\begin{figure}
    \centering
    \includegraphics[scale=0.34]{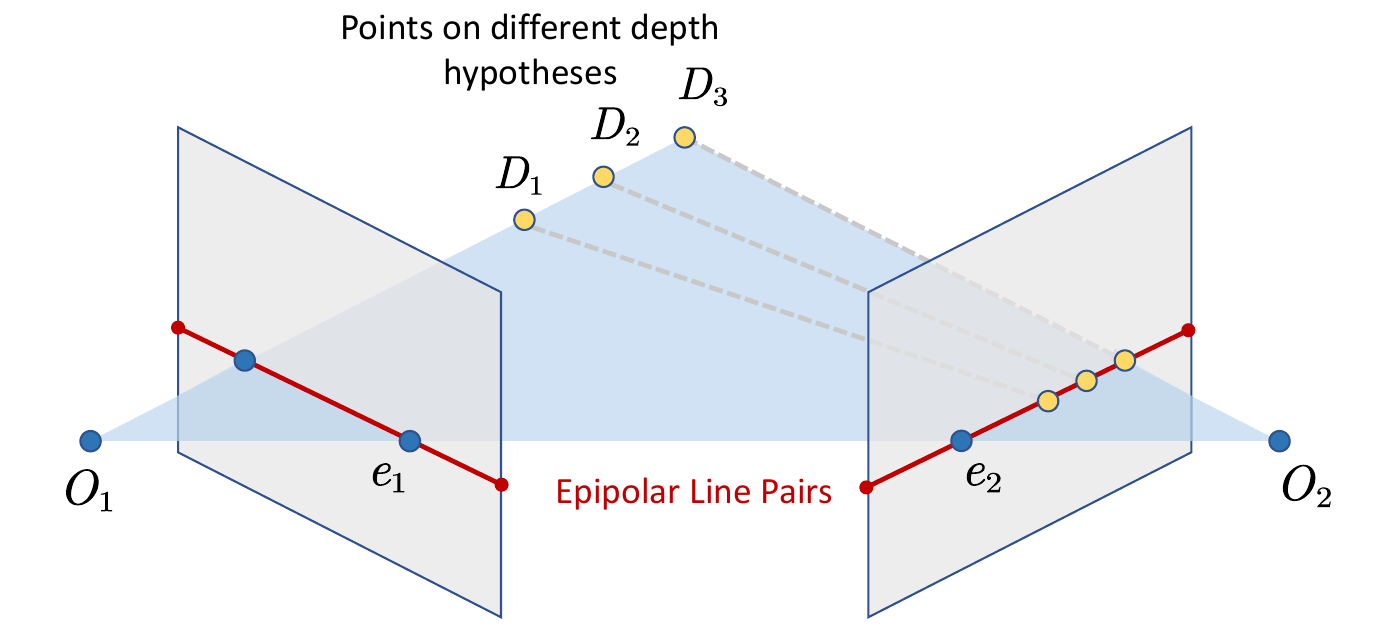}
    \caption{
    \textbf{Epipolar Geometry.} A pixel with different depth hypotheses in the reference image can be projected to the same epipolar line in the source image, termed point-to-line. The plane that passes through two camera centers and a 3-D point produces an epipolar line when intersecting with an image plane. The two epipolar lines of two images form an epipolar line pair, termed line-to-line.
    }
    \label{epipolar}
\end{figure}

According to the epipolar geometry theory, the actual point corresponding to a pixel in one view can only be projected on the epipolar line in the other view. Hence, when searching the corresponding points, the 2D search space can be reduced into a 1D line. This reduces the computation and eliminates background interference.
Naturally, each source image can form a stereo matching problem with the reference image. As shown in Fig.~\ref{epipolar}, points with different depth hypotheses will be projected onto the epipolar line in the other image. Therefore, the feature encoding in MVS can be viewed as describing different points on the epipolar line. 
%

%
Since the points to be distinguished lie on the same line, inspired by this epipolar constraint, we propose a line-to-point feature aggregation strategy. For each point, the information from the corresponding epipolar line is used to describe it.
A similar line-based feature aggregation has been adopted by CSwin~\cite{dong2022cswin} for vision backbones, which is proved to be efficient and effective. We find that for MVS, with epipolar geometry as a theoretical basis, using epipolar lines as the source of non-local features can efficiently achieve high-quality representation, as shown in Fig.~\ref{fig1} (b).

%

Specifically, we propose an epipolar line-guided non-local mechanism, which only fetches features from the corresponding epipolar lines. First, we decompose the reference and source feature map into pixel groups; each group shares identical epipolar line pairs. This process is processed with an optimized search algorithm. For each point in the reference image, we calculate its corresponding epipolar line in the source image. Then, points with approximate epipolar line parameters are clustered into the same group. With this step, all the pixels in the source and reference features can find their epipolar line pairs. Then, as shown in Fig.~\ref{fig1} (a), we perform non-local feature aggregation within the source epipolar line and across the epipolar line pairs. We also add a local augmentation module to mitigate the discontinuities of epipolar line partition. We instantiate this epipolar line-guided non-local mechanism with a Transformer model termed Epipolar Transformer (ET).


To demonstrate the effectiveness of the proposed modules, we present a coarse-to-fine framework termed ET-MVSNet, which applies the Epipolar Transformer (ET) to the feature encoding part of recent works~\cite{gu2020cascade,wang2022mvster}. We conduct extensive experiments to demonstrate the effect of our method. Thanks to Epipolar Transformer (ET), the depth estimation and 3-D representations have greatly improved and achieved state-of-the-art results on both the DTU~\cite{dtu} and the Tanks and Temple~\cite{tanks} benchmark. Besides, compared with the prevailing global attention that mines global context, Epipolar Transformer (ET) shows significantly better performance and efficiency.

\vspace{5pt}

Our main contributions can be summarized as follows:
\begin{itemize}
    \item We propose an epipolar line feature aggregation strategy for the representation learning of MVS.

    \item We devise an optimized epipolar pair search algorithm, and an Epipolar Transformer (ET) to enable non-local feature augmentation on epipolar lines.

    \item Our method achieves state-of-the-art performance on both the DTU dataset and Tanks-and-Temples benchmark.

\end{itemize}



\section{Related Work}
\noindent \textbf{Learning-based MVS.}
The core of MVS is a feature-matching problem, where representation and similarity metrics play critical roles. The performance of traditional MVS is limited due to poor representation ability, especially in weak-texture or repetitive areas. Powered by the strong representation of neural networks, the learning-based MVS is first proposed by MVSNet~\cite{yao2018mvsnet} and quickly becomes the main stream due to better robustness and performance. MVSNet proposes an end-to-end pipeline for depth estimation with a post-processing procedure for fusion. Subsequent work improves the learning-based MVS from different aspects. \textit{e.g.}, reducing memory and computation overhead with RNN-based approaches~\cite{wei2021aa,yan2020dense,yao2019recurrent} and coarse-to-fine paradigms~\cite{cheng2020deep,gu2020cascade,yang2020cost,zhang2020visibility}, optimizing the cost aggregation by adaptively re-weighting different views~\cite{luo2019p,yi2020pyramid}. An important line is to enhance the feature representation with a non-local feature aggregation mechanism, \textit{e.g.}, Deformable Convolution Network~\cite{dai2017deformable}. However, existing non-local techniques suffer from expensive computational overheads attributable to multi-image inputs. In this paper, we focus on the characteristics of the MVS task and propose a novel and efficient feature enhancement method.

\noindent \textbf{Feature Representation in MVS.}
To obtain descriptive representation for matching, Dai \textit{et al.}~\cite{dai2017deformable} utilizes deformable convolution~\cite{dai2017deformable} to capture instance cues for splitting foreground and background, such that the context of them would not interact and the feature representations are not sensitive to the perspective changing. Transformer~\cite{vaswani2017attention} was first introduced into the MVS pipeline by TransMVSNet~\cite{ding2022transmvsnet} for its global receptive field and dynamic feature aggregation ability. TransMVSNet applies intra- and inter-image attention to transfer context and uses a linear Transformer~\cite{katharopoulos2020transformers} to reduce computational cost. MVSTER~\cite{wang2022mvster} employs a parameter-free attention mechanism to re-weight the aggregated contribution of different pixels. The success of the attention mechanism highlights the significance of learning robust feature representations, as the essence of MVS lies in feature matching. Although more advanced feature aggregation approaches such as attention are introduced for MVS, the efficiency decreases significantly. The existing Transformer-based MVS methods adopt efficient attention operators, while proper constraints on the feature aggregation remain unstudied. In this paper, inspired by epipolar geometry constraints, we propose to perform non-local feature augmentation only on the epipolar pairs.

\section{Preliminaries}
\noindent \textbf{Learning-based MVS.}
Most learning-based methods follow a cost volume-based pipeline first introduced by MVSNet~\cite{yao2018mvsnet}. Given a reference image and $N-1$ source images, learning-based methods aim to recover the depth map of the reference image. The pipeline consists of four steps. The first step is the feature encoding where images are projected in feature representation. In the second step, the cost volume is constructed by feature matching. Specifically, according to a set of predefined depth hypotheses, each source feature map is warped into reference camera frustum via differentiable homography to measure their similarity and obtain feature volumes. These $N-1$ feature volumes are then aggregated to construct a 3D cost volume that encodes the cost of each depth hypothesis for the reference image. After that, a 3D CNN can be applied to the cost volume for regularization and obtain the probability volume to infer depths. As mentioned above, the basic idea of depth estimation is to perform point matching to estimate final depths based on predefined depth hypotheses, where the feature representation capacity directly affects the accuracy of matching. Therefore, we aim to propose an effective and efficient information aggregation strategy to capture non-local context and enhance feature representation.




\noindent \textbf{Epipolar Constraint}.
\label{Epipolar Constraint} The epipolar constraint is a common geometric constraint in stereo matching. Given a pair of stereo images, stereo matching aims to find the corresponding points between the two images. As shown in Fig.~\ref{epipolar}, for a given point in one image, its corresponding point in the other image must lie on a specific line, which is called the epipolar line. Geometrically, an epipolar line is an intersection between the image plane and the plane passing through the two camera centers and the corresponding 3D point. In addition, epipolar lines occur in pairs, meaning that any point on one epipolar line corresponds to the same epipolar line in the other image.

With this constraints, the search space of the stereo matching~\cite{cox1996maximum,bleyer2011patchmatch,fusiello2000compact,hirschmuller2007stereo,scharstein2002taxonomy}, can be reduced from the whole image plane into a single line. MVS can be viewed as a multiple stereo matching problems. Points with different depth hypotheses in the reference view will be projected on the epipolar line in the source images. Hence, the feature encoding can be viewed as describing different points which are on the same line. 


\begin{figure}
    \centering
    \includegraphics[scale=0.27]{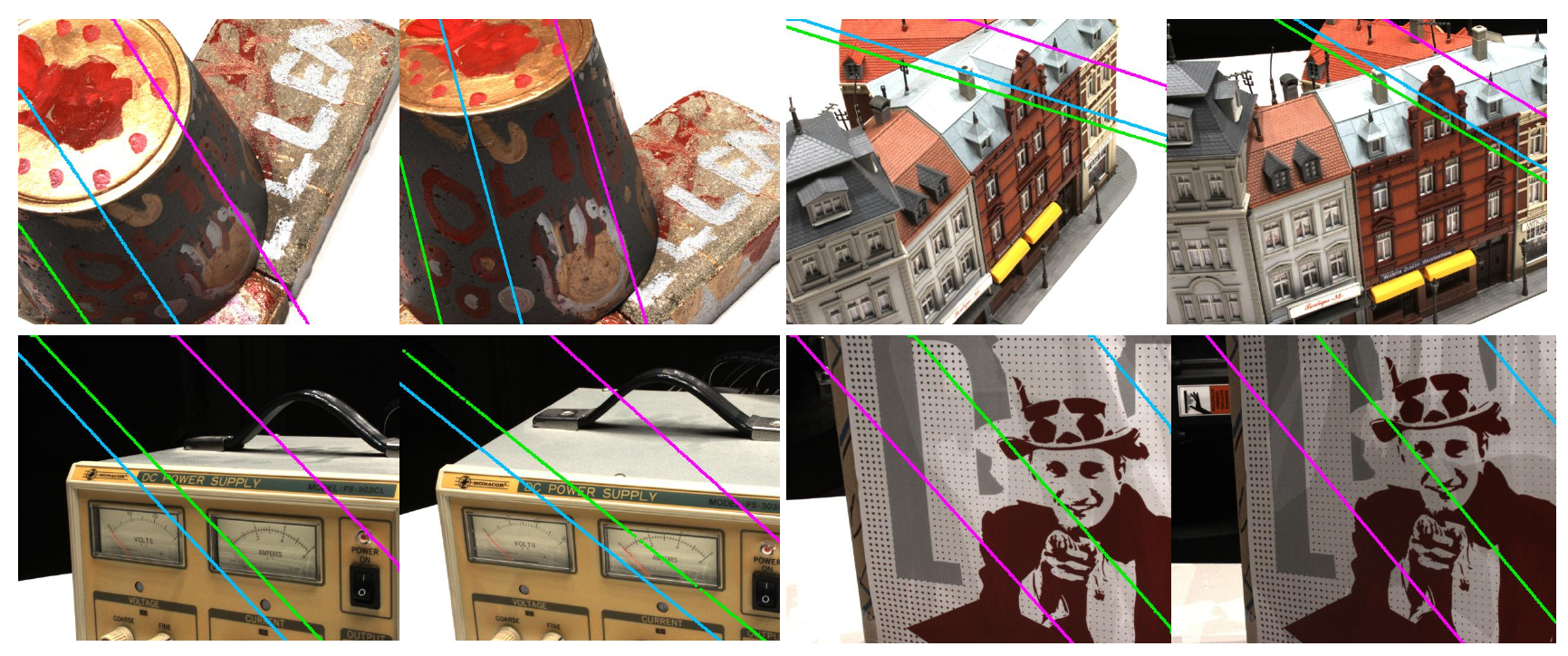}
    \caption{\textbf{Visualization of the epipolar pairs.} The same color represents the corresponding epipolar lines.}
    \label{fig2}
    \vspace{-10pt}
\end{figure}
\section{ET-MVSNet}
\begin{figure*}
    \centering
    \includegraphics[width=0.95\textwidth]{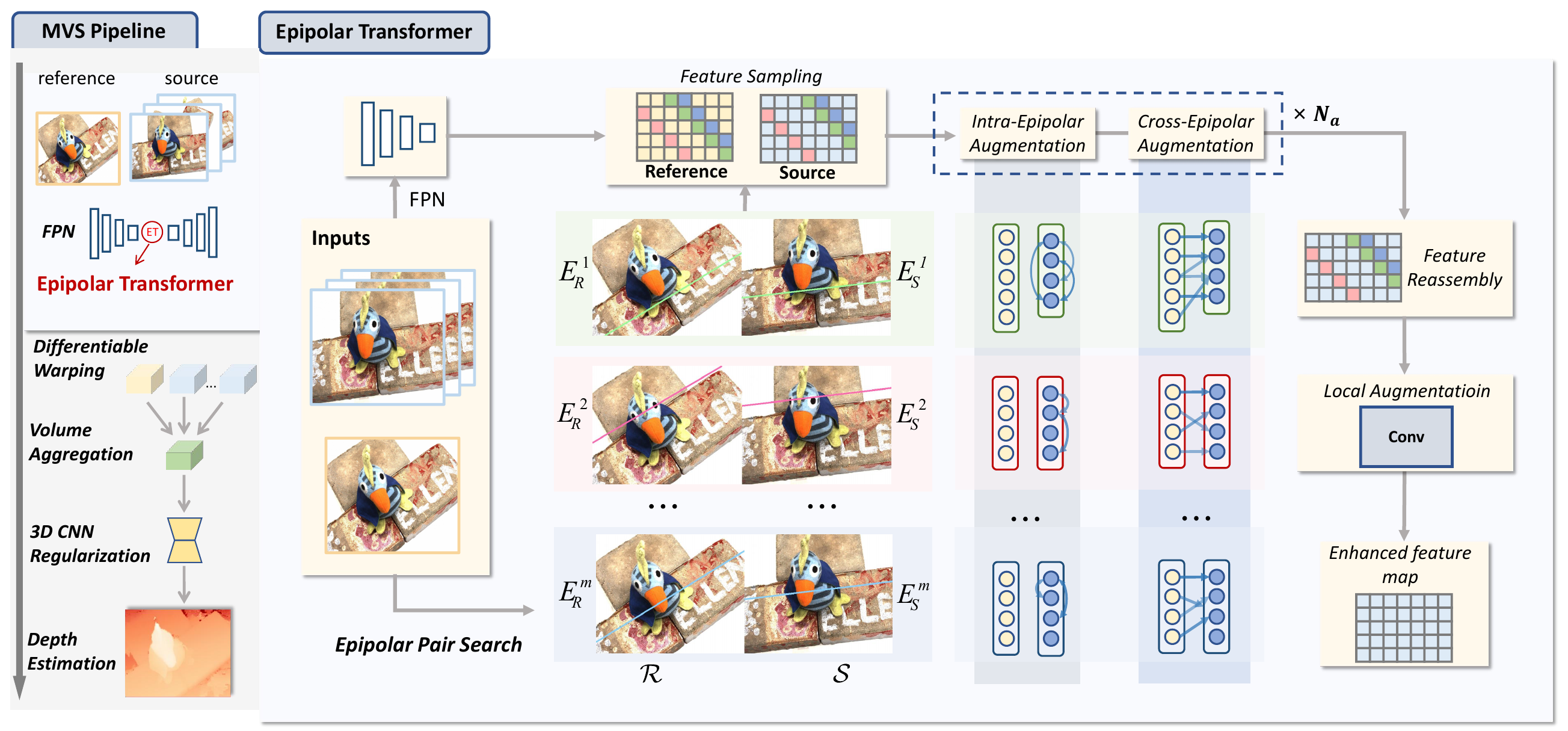}
    \caption{\textbf{ET-MVSNet architecture.} We apply the Epipolar Transformer (ET) to common learning-based MVS frameworks by integrating it into the FPN. Firstly, the epipolar pairs are searched via our algorithm and the epipolar feature sequences are obtained by sampling the coarsest feature map of the FPN. The sequence is then fed into the Intra-Epipolar Augmentation (IEA) and Cross-Epipolar Augmentation (CEA) blocks, which can be stacked $N_a$ layers. In our implementation, we set $N_a=1$. The resulting sequences are mapped back to a feature map, followed by the Local Augmentation (LA) module to smooth the enhanced feature map. Finally, the enhanced feature map is sent to the upsampling layers of the FPN, followed by cost volume construction and regularization.}
    \label{fig4}
\end{figure*}

The overall architecture of our ET-MVSNet is illustrated in Fig.~\ref{fig4}. The proposed Epipolar Transformer (ET) module is integrated into the Feature Pyramid Network (FPN)~\cite{lin2017feature}. To perform non-local feature aggregation on epipolar lines, we first search the epipolar line pairs between reference and source images. With the searched line pairs, the original feature maps are decomposed into different epipolar line pairs via sampling. Then, the Intra-Epipolar Augmentation (IEA) and the Cross-Epipolar Augmentation (CEA) modules transfer non-local context among these epipolar line pairs.


\begin{algorithm}[t]
\label{algorithm 1}
    \SetAlgoLined
    \KwIn{Camera parameters; image size $(H,W)$; parameters matrices  $\mathbf{K},\mathbf{B}$ of $k$s,$b$s; seach threshold $\delta$. }
    \KwOut{Epipolar pairs}
        
    \For{$\left(x,y\right) = \left(0,0\right)$ \KwTo $\left(W-1,H-1\right)$}
    {
        calculate slope $\mathbf{K}^{x,y}$ and bias $\mathbf{B}^{x,y}$ \\
        
    }

    \For {$k$ \bf{in} $set(Quantify(\mathbf{K}))$}
    {
        \For {$b$ \bf{in} $set(Quantify(\mathbf{B}))$}
        {
             \For{$\left(x,y\right) = \left(0,0\right)$ \KwTo $\left(W-1,H-1\right)$}
             {
                epipoline\_ref : $\mathbf{K}^{x,y} = k$ and $\mathbf{B}^{x,y} = b$ \\
                epipoline\_src : $\frac{|y-kx-b|}{\sqrt{1+k^{2}}} < \delta$
             }
         }
    }
    
    \caption{Epipolar Pair Search}
    
\end{algorithm}

\subsection{Epipolar Pair Search}
\label{Epipolar Pair Searching}



The key component of MVS is to match pixels to the most suitable depth hypothesis from a set of pre-defined hypotheses through feature matching. If a hypothesis is close to the actual depth of the ground, the features of the pixel pairs are supposed to be similar, indicating that the discrimination of the hypothesis relies heavily on feature representation. Since the feature volume used for matching naturally lies on the epipolar line, using epipolar lines as the source of non-local features can efficiently achieve high-quality representation, benefiting in distinguishing different hypotheses. It should be noted that epipolar lines exist in both the reference view and the source view and appear in pairs due to geometric constraints. It suggests the pixels on the same epipolar line share the same source of non-local features and the process of aggregating features for these pixels is the same. A parallel aggregation would improve the efficiency. To realize the parallel process, pixels located on the same epipolar line pair are required to be pre-searched and we propose an epipolar-pair search algorithm for it.

As shown in Algorithm~\ref{algorithm 1}, the searching of epipolar pairs consists of two steps: point-to-line search and epipolar line matching. In the first step, we compute the epipolar line parameters for each pixel in the reference view. Then, we aggregate the pixels in the reference view into different clusters; pixels in the same cluster share the same epipolar line. Hence, after the epipolar pair search, pixels in the reference and source images are partitioned into epipolar line pairs. 

 Since a one-to-many view matching can be decomposed into multiple one-to-one view matching, for easy understanding, here we illustrate the epipolar pair search with one source view. Given a pixel $\mathbf{p}_r$ in the reference view, the corresponding pixel $\mathbf{p}_s$ in the source view is
\begin{equation}
\label{eq:equation 1}
\mathbf{p}_s(d) = \mathbf{K}_s [\mathbf{R}(\mathbf{K}_r^{-1}\mathbf{p}_rd)+\mathbf{t}]\,,
\end{equation}
where $d$ denotes the depth. $\mathbf{R}$ and $\mathbf{t}$ denote the rotation and translation between the reference and the source view. $\mathbf{K}_r$ and $\mathbf{K}_s$ denote the intrinsic matrices of the reference view and source view, respectively. 
Hence, the coordinate of $\textbf{p}_s(d)$ can be computed by
\begin{equation}
x_s(d)  = \frac{a_1 d + b_1}{a_3 d + b_3}\,,
y_s(d)  = \frac{a_2 d + b_2}{a_3 d + b_3}\,,
\label{eq:coordinates}
\end{equation}
where $\{a_i\}_{i=1}^3$ and $\{b_i\}_{i=1}^3$ are constants associated with the camera parameters and the coordinate of $\mathbf{p}_r$~(refer to the Supplementary Materials for more details). Then, we can eliminate the depth $d$ to obtain the standard equation $y_s(d)=kx_s(d)+b$ of the epipolar line, which is formulated as:

\begin{equation}
\label{equation 3}
\left\{
\begin{aligned}
k &= \frac{\Delta y_s(d)}{\Delta x_s(d)} = \frac{a_2 b_3 -a_3 b_2}{a_1 b_3 -a_3 b_1} \\
b &= y_s(0) - k x_s(0) = \frac{b_2}{b_3} - k \frac{b_1}{b_3}
\end{aligned}
\right.\,.
\end{equation}
Specifically, when ${\Delta x_s(d)}\rightarrow 0$, we use the equation $x_d(s) = k^{'}y_s(d) + b^{'}$. 
As $k, b$ are independent of the $d$ and are expressed by constants associated with the coordinate of $\mathbf{p}_r$, if two pixels are on the same epipolar line, the $k$s and $b$s of them would be theoretically the same. However, in practice, due to the discrete coordinates of pixels, the $k$s and $b$s calculated by different pixels around the epipolar line may vary in a tiny range, such that directly grouping the pixels with identical $k$s or $b$s will lead to over-splitting.

To alleviate the over-splitting, we quantify $k$s and $b$s by rounding. Therefore, pixels with approximate parameters will be grouped into the same epipolar line. Have finished the grouping of the reference view. Then, we search for the corresponding epipolar line in the source view. 
The $k$s and $b$s represent the corresponding epipolar lines on the source view, as the coordinates of pixels around the epipolar line will meet \cref{equation 3}. For the source image, whether a pixel is on the epipolar line expressed by $k$ and $b$ is determined by calculating the distance of them.

By dividing pixels into their corresponding epipolar pairs, both the reference and the source feature maps are decomposed into sets of feature sequences. Specifically, supposing that $m$ pairs of epipolar lines are obtained, we define the reference and source feature sets as $E_{\mathcal{R}}$ and $E_{\mathcal{S}}$ which are formulated as:
\begin{equation}
    \label{equation 4}
    \begin{aligned}
    E_{\mathcal{R}} = \{E_R^1, E_R^2, ..., E_R^m\},\  E_{\mathcal{S}} = \{E_S^1, E_S^2, ..., E_S^m\}\,.
    \end{aligned}
\end{equation}
$E_R^i$ and $E_S^i$ are feature sequences with the shape of $n\times c$, where $n$ denotes the number of pixels in the corresponding epipolar line, and $c$ denotes the feature dimension. We visualize the searched epipolar line pairs in Fig.~\ref{fig2}.

\subsection{Intra- and Cross-Epipolar Augmentation}
\label{Epipolar Transformer}

With the epipolar line pairs $E_{\mathcal{R}}$ and $E_{\mathcal{S}}$, we then perform the non-local feature augmentation. Since points with different depth hypotheses fall on an epipolar line, the matching process is to distinguish a series of points lying on the same line. Inspired by the point-to-line searching strategy in stereo matching, we propose a line-to-point non-local augmentation: each pixel in the reference image only attends to its corresponding epipolar line pairs. 

As shown in Fig.~\ref{fig4}, to describe the pixel with non-local information in the source image, an Intra-Epipolar Augmentation (IEA) module is devised based on self-attention. A Cross-Epipolar Augmentation (CEA) module then propagates the information on reference epipolar line $E_{\mathcal{R}}$ into the source epipolar line $E_{\mathcal{S}}$ with cross-attention. 



\noindent \textbf{Intra-Epipolar Augmentation (IEA)}.
IEA utilizes the self-attention within an epipolar line to aggregate non-local structural information, which can generate descriptive feature representations for difficult regions, \textit{e.g.}, weak-texture areas.
For each epipolar line $E_S^i$ in the $E_{\mathcal{S}}$, the augmentation process is defined as:
\begin{equation}
\label{equation 5}
    \begin{aligned}
    E_S^i=\mathrm{MHSA}\left(E_S^i \right) \,\,+\,\,E_S^i\,,
    \end{aligned}
\end{equation}
where $\mathrm{MHSA}(x)$ refers to the multi-head self attention~\cite{vaswani2017attention} that takes a sequence $x$ as input. 

\noindent \textbf{Cross-Epipolar Augmentation (CEA)}.
Due to the varying viewpoints, the potential perspective transformation renders challenges for the same semantic pixels with different geometries. We alleviate this by transferring information across the epipolar lines.
Specifically, we use a cross-attention module to propagate the information from reference line $E_{\mathcal{R}}$ into the $E_{\mathcal{S}}$, denoted by CEA. In CEA, $E_S^i$ is first processed by a cross-attention layer:
\begin{equation}
    E_S^i=\mathrm{MHCA}\left(E_S^i, E_R^i, E_R^i\right) \,\,+\,\,E_S^i\,,
\end{equation}
where $\mathrm{MHCA}(q,k,v)$ refers to the multi-head cross attention~\cite{vaswani2017attention} layer with $q$, $k$ and $v$ as the input query, key and value sequences. Specifically, we add a feed-forward network as in the standard Transformer block~\cite{vaswani2017attention} after the cross-attention layer. The block of IEA and CEA can be stacked for feature augmentation.

\vspace{5pt}
\noindent{\textbf{Local Augmentation (LA)}.}
Although the non-local feature augmentation within epipolar lines is efficient and proved to be effective in our experiments.
We find that the enhanced feature map contains some holes where pixels that are located outside the common view of two images or are not detected by the algorithm due to the quantification error caused by the discrete nature of pixels, which can lead to discontinuities in feature representation and is unfriendly for matching.
To remedy this issue, we use an additional convolution layer after the IEA and CEA block to re-aggregate the local context for filling feature holes and smoothing the augmented feature.

\begin{figure*}[t]
    \centering
    \includegraphics[width=0.99\textwidth]{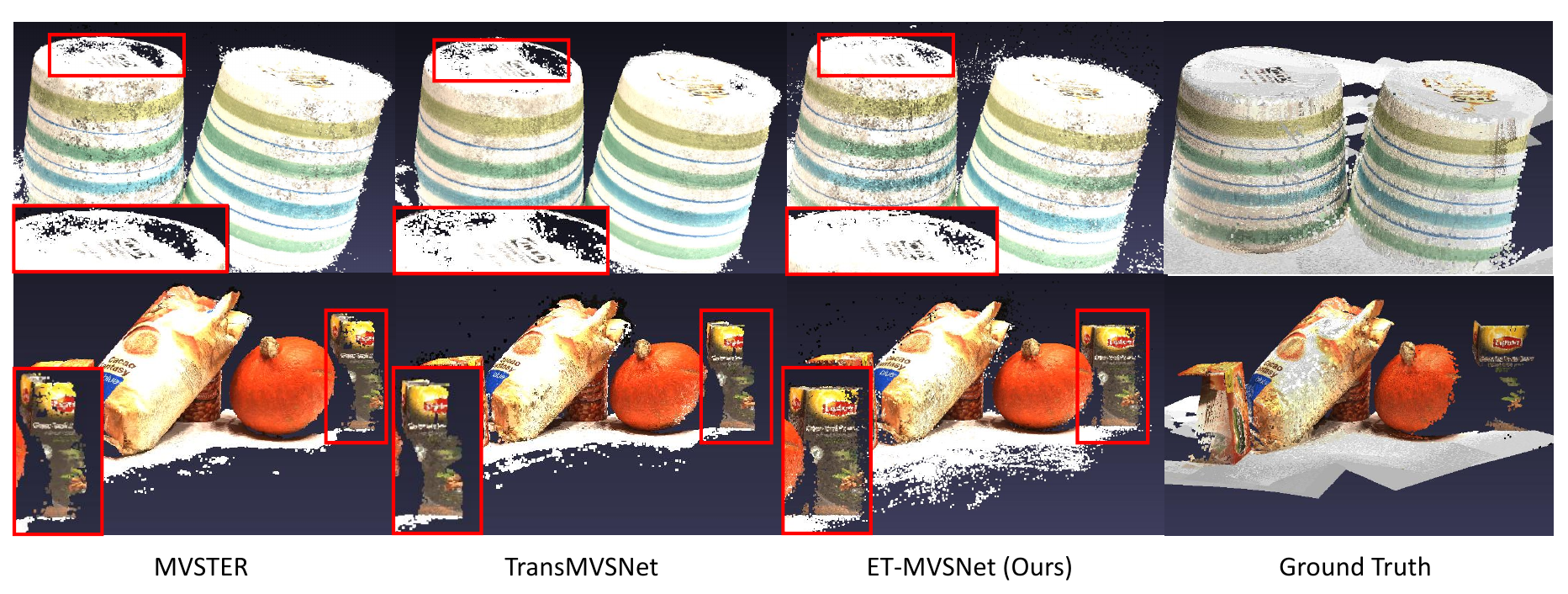}
    \caption{\textbf{Comparison of reconstructed results with state-of-the-art methods~\cite{wang2022mvster,ding2022transmvsnet} on DTU evalution set~\cite{dtu}.}}
  
    \label{dtu_vis}
    \vspace{-10pt}
\end{figure*}

\subsection{Implementation Details}
\label{Implementation Details}
Inherited from MVSTER~\cite{wang2022mvster}, we use a four-stage depth estimation with 8, 8, 4, and 4 depth hypotheses, and adopt the inverse depth sampling. To simplify the training, the original monocular depth estimation auxiliary branch from MVSTER is removed. We integrate the proposed Epipolar Transformer (ET) into the FPN processing the coarsest feature map with a downsampling rate of $8$ to save computation costs. 

We use the cross-entropy loss for all stages. The total loss is as follows:

\begin{equation}
\label{equation 7}
Loss = \sum_{k=1}^{N} \lambda ^k  L^k
\end{equation}
where $N$ refers to the number of stages, $L^k$ refers to the loss at the $k^{th}$ stage and $\lambda ^k$ refers to loss weight of the $k^{th}$ stage. Specifically, $N=4$ and $\lambda ^1 = ... =\lambda ^N = 1$ in our implementation.

\section{Experiments}
In this section, we first introduce datasets, metrics, and experiment settings in Sec.~\ref{Settings}, and then present the benchmark performance in Sec.~\ref{Benchmark Performance}. Additionally, we perform ablation studies to demonstrate the effectiveness of each component and the efficiency of our proposed feature aggregation approach in Sec.~\ref{Ablation Study}.

\subsection{Settings}
\label{Settings}

\noindent \textbf{DTU dataset.} DTU is an indoor dataset collected under well-controlled laboratory conditions with a fixed camera trajectory. The DTU dataset contains $128$ different scenes with $49$ or $64$ views under $7$ different illumination conditions. We adopt the same data partition as MVSNet~\cite{yao2018mvsnet} for a fair comparison. We take 5 images and follow the dynamic checking strategy~\cite{yan2020dense} for depth filtering and fusion. Accuracy and completeness are used as the evaluation metrics. In addition, an ``overall'' metric which averages the accuracy and completeness is reported.

\noindent \textbf{Tanks and Temples benchmark.} ``Tanks and Temples'' is a public benchmark consisting of $14$ outdoor realistic scenes, including an intermediate subset of $8$ scenes and an advanced subset of $6$ scenes. We take 11 images and follow the dynamic checking strategy~\cite{yan2020dense} for depth filtering and fusion. Note that the ground truth for the Tanks and Temples benchmark is hidden, all results are evaluated on an official website. We use F-score as the evaluation metric, which combines both precision and recall. Refer to the Supplementary Materials for more details of depth fusion.

Following common practices~\cite{gu2020cascade,ding2022transmvsnet,unimvs,wang2022mvster}, before evaluating on Tanks and Temples benchmark, we finetune our model on the BlendedMVS dataset~\cite{yao2020blendedmvs} for the adaptation to the real-world scenes. We use the original image resolution $576 \times 768$ and the number of images equals $7$.

\begin{table*}
\centering
\setlength{\tabcolsep}{3pt}
\resizebox{1\linewidth}{!}{
\begin{tabular}{@{}l|ccccccccc|ccccccc@{}}
\toprule
\multirow{2}{*}{Method}&\multirow{2}{*}    & \multicolumn{8}{c|}{Intermediate} & \multicolumn{6}{c}{Advanced}\\ 
 & \textbf{Mean} &Fam. &Fra. &Hor. &Lig. &M60 &Pan. &Pla. &Tra.  & \textbf{Mean} &Aud. &Bal. &Cou. &Mus. &Pal. &Tem.\\
\midrule

COLMAP~\cite{colmap} & 42.14 & 50.41 & 22.25 & 26.63 & 56.43 & 44.83 & 46.97 & 48.53 & 42.04 & 27.24 & 16.02 & 25.23 & 34.70 & 41.51 & 18.05 & 27.94	\\

Point-MVSNet~\cite{chen2019point} & 48.27 &61.79 &41.15 &34.20 &50.79 &51.97 &50.85 &52.38 &43.06 &- &- &- &- &- &- &- \\

R-MVSNet~\cite{yao2019recurrent} &  50.55 & 73.01 & 54.46 & 43.42 & 43.88 & 46.80 & 46.69 & 50.87 & 45.25 & 29.55 & 19.49 & 31.45 & 29.99 & 42.31 & 22.94 & 31.10 \\
PatchmatchNet~\cite{wang2021patchmatchnet} & 53.15 &66.99 &52.64 &43.24 &54.87 &52.87 &49.54 &54.21 &50.81 &32.31 &23.69 &37.73 &30.04 &41.80 &28.31 &32.29 \\

P-MVSNet~\cite{luo2019p}  & 55.62 &70.04 &44.64 &40.22 &\underline{65.20} &55.08 &55.17 &60.37& 54.29 &- &- &- &- &- &- &- \\

CasMVSNet~\cite{gu2020cascade} & 56.84 &76.37 &58.45 &46.26 &55.81 &56.11 &54.06 &58.18 &49.51  &31.12 &19.81 &38.46 &29.10 &43.87 &27.36 &28.11 \\ 

MVSTER~\cite{wang2022mvster} & 60.92  &80.21	&63.51	&52.30	&61.38	&61.47	&58.16	&58.98	&51.38  &37.53 &26.68 &42.14 &35.65 &49.37 &32.16 &\underline{39.19} \\

GBi-Net~\cite{mi2022generalized} & 61.42 &79.77 &\underline{67.69} &51.81 &61.25 &60.37 &55.87 &60.67 &53.89 &37.32 &\textbf{29.77} &42.12 &36.30 &47.69 &31.11 &36.93  \\

AA-RMVSNet~\cite{wei2021aa} & 61.51 &77.77 &59.53 &51.53 &64.02 &64.05 &59.47 &60.85 &55.50 &33.53 &20.96 &40.15 &32.05 &46.01 &29.28 &32.71 \\

EPP-MVSNet~\cite{ma2021epp}  & 61.68 &77.86 &60.54 &52.96 &62.33 &61.69 &60.34 &\textbf{62.44} &55.30 &35.72 &21.28 &39.74 &35.34 &49.21 &30.00 &38.75 \\

TransMVSNet~\cite{ding2022transmvsnet} & 63.52 &80.92 &65.83 &\underline{56.94} &62.54 &63.06 &60.00 &60.20 &\underline{58.67} &37.00 &24.84 &\underline{44.59} &34.77 &46.49 &\underline{34.69} &36.62  \\

UniMVSNet~\cite{unimvs} & \underline{64.36} &\underline{81.20} &66.43 &53.11 &63.46 &\textbf{66.09} &\textbf{64.84} & \underline{62.23} &57.53 &\underline{38.96} &28.33 &44.36 &\textbf{39.74} &\textbf{52.89} &33.80 &34.63 \\

\midrule

ET-MVSNet (Ours)  & \textbf{65.49} & \textbf{81.65} & \textbf{68.79} & \textbf{59.46} & \textbf{65.72} & \underline{64.22} & \underline{64.03} & 61.23 & \textbf{58.79} & \textbf{40.41} & \underline{28.86} & \textbf{45.18} & \underline{38.66} & \underline{51.10} & \textbf{35.39} & \textbf{43.23}\\

\bottomrule
\end{tabular}}
\vspace{2pt}
\caption{\textbf{Quantitative results on Tanks and Temples benchmark.} The metric is F-score and ``Mean" refers to the average F-score of all scenes (\textbf{higher is better}). The best and the second-best results are in \textbf{bold} and \underline{underlined}, respectively.}
\label{TT table}
\end{table*}

  

\begin{table}\Huge
\setlength{\tabcolsep}{3pt}

\resizebox{1.\linewidth}{!}{
\begin{tabular}{lccc}
\toprule
Method & ACC.(mm) $\downarrow$ & Comp.(mm) $\downarrow$ & Overall(mm) $ \downarrow$ \\
\midrule
Gipuma~\cite{gipuma}         & \textbf{0.283} & 0.873 & 0.578 \\
COLMAP~\cite{colmap}         & 0.400 & 0.664 & 0.532 \\
SurfaceNet~\cite{surfacenet}     & 0.450 & 1.040 & 0.745 \\
P-MVSNet~\cite{luo2019p}       & 0.406 & 0.434 & 0.420 \\
R-MVSNet~\cite{yao2019recurrent}       & 0.383 & 0.452 & 0.417 \\
Point-MVSNet~\cite{chen2019point}   & 0.342 & 0.411 & 0.376 \\
AA-RMVSNet~\cite{wei2021aa}     & 0.376 & 0.339 & 0.357 \\
EPP-MVSNet~\cite{ma2021epp}     & 0.413 & 0.296 & 0.355 \\
PatchmatchNet~\cite{wang2021patchmatchnet}  & 0.427 & 0.277 & 0.352 \\
CasMVSNet~\cite{gu2020cascade}      & 0.325 & 0.385 & 0.355 \\
MVSTER~\cite{wang2022mvster}         & 0.350 & 0.276 & 0.313 \\
UniMVSNet~\cite{unimvs}      & 0.352 & 0.278 & 0.315 \\
TransMVSNet~\cite{ding2022transmvsnet}    & \underline{0.321} & 0.289 & 0.305 \\
GBiNet~\cite{mi2022generalized}         & 0.327 & \underline{0.268} & \underline{0.298} \\
\midrule
ET-MVSNet (Ours) & 0.329 & \textbf{0.253} & \textbf{0.291}        \\
\bottomrule
\end{tabular}}
\vspace{2pt}
\caption{\textbf{Quantitative point cloud results on DTU evaluation set.} (\textbf{lower is better}). The best and the second-best results are in \textbf{bold} and \underline{underlined}, respectively.}
\label{DTU Table}
\vspace{-11pt}
\end{table}

\subsection{Benchmark Performance}
\label{Benchmark Performance}

\noindent \textbf{Evaluation on DTU}. We report standard metrics by using the official evaluation script. The quantitative results of the DTU evaluation set are shown in Table~\ref{DTU Table}. Compared with other state-of-the-art methods, our method can reconstruct denser point clouds, which leads to advantageous completeness. With improved completeness, our method retains comparable accuracy and obtains the best overall metric. Fig.~\ref{dtu_vis} shows qualitative results on the DTU evaluation set. Our method can reconstruct more detailed point clouds in challenging areas, such as weakly textured surfaces.

\noindent \textbf{Evaluation on Tanks and Temples}. To verify the generalization ability of our method, we evaluate our method on the Tanks and Temples benchmark. The quantitative results of intermediate and advanced sets are shown in Table~\ref{TT table}. Our method outperforms the other MVS approaches on both the intermediate and the advanced sets, which demonstrates the strong generalization ability of our method. Refer to Supplementary Materials for qualitative results.

\subsection{Ablation Study}
\label{Ablation Study}
We conduct a series of ablation studies to analyze the effectiveness and efficiency of the proposed module, including different components in the Epipolar Transformer (ET) and different feature aggregation operators. All the ablation is conducted based on the MVSTER\cite{wang2022mvster}. We conduct experiments on the DTU dataset with metrics in~\cref{DTU Table} and an additional depth error. The depth error refers to the average absolute values of the differences between predicted depths and ground-truth depths, whose resolution is  $512 \times 640$.  

\begin{table}\large
\centering
\setlength{\tabcolsep}{8pt}
\renewcommand\arraystretch {1.05}
\resizebox{1.\linewidth}{!}{
\begin{tabular}{@{}ccc|ccccc@{}}
\toprule
\multicolumn{3}{@{}l|}{Settings}  & \multirow{2}{*}{ACC.$\downarrow$} & \multirow{2}{*}{Comp.$\downarrow$} & \multirow{2}{*}{Overall$\downarrow$} & \multirow{2}{*}{Depth Error$\downarrow$} \\
 IEA & CEA & LA & & &  \\
\midrule
 &  & &  0.351 & 0.284 & 0.318 & 6.355 \\
 \checkmark & \checkmark  &  &  0.331 & 0.263 & 0.297 &  5.843 \\
 \checkmark & &  \checkmark &   0.327 &  0.261&0.294  &  6.021 \\
& \checkmark  &  \checkmark&  \textbf{0.326} &  0.262 & 0.294 &5.982   \\
 \checkmark & \checkmark & \checkmark &  0.329 & \textbf{0.253} & \textbf{0.291} & \textbf{5.754}\\
\bottomrule
\end{tabular}
}
\vspace{2pt}
\caption{\textbf{Ablations on the DTU evaluation set.}}
\label{Benefits of ET}
\vspace{-6pt}
\end{table}

\noindent \textbf{Benefits of ET}
As shown in Table~\ref{Benefits of ET}, we ablate different components in the proposed ET. First, comparing row 1 with row 2, including the ICE and CEA to mining non-local context along the epipolar line decreases the depth error by $8\%$, leading to a $+6\%$ improvement in the quality~(Overall) of 3-D reconstruction. Comparing row 2 with row 5, adding the LA to address the discontinuities on feature representation further yields a $+7\%$ improvement in 3-D reconstruction accuracy. Comparing row 3 and row 4 with row 5, only using IEA or CEA for intra-line and inter-line augmentation will slightly decrease the overall metrics, but can still obtain better performance than no non-local augmentation. The combination of ICE, CEA, and LA achieves the lowest depth error and the best 3-D reconstruction results.

\begin{figure}
    \centering
    \includegraphics[width=0.45\textwidth]{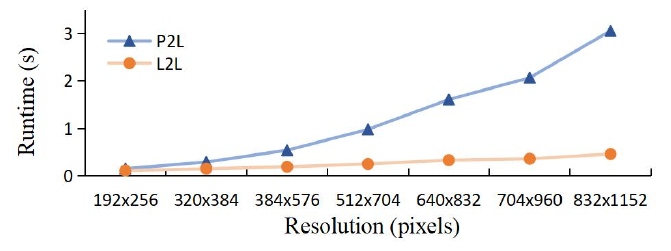}
    \caption{\textbf{Different information interaction approaches for images of different resolutions.} ``P2L" refers to the interaction between a point and the corresponding epipolar line, and ``L2L" refers to the interaction between epipolar pairs.}
    \label{fig5}
    \vspace{-2pt}
\end{figure}

\noindent \textbf{Epipolar line search algorithm}. 
Here we discuss the efficiency of different implementations for the line search algorithm. As illustrated in Sec.~\ref{Epipolar Pair Searching}, we pre-calculate the epipolar lines for each pixel, and group pixels with approximate epipolar lines into the same cluster. We define this implementation as line-to-line, as the outputs are pairs of lines. A simple alternative can be a point-to-line implementation: 
calculate the line parameters for each pixel in the source view individually.

We compare two different implementations in Table~\ref{Efficiency of line-to-line}. With comparable performance, point-to-line increases the computation overhead (MACs) by $50$ times and takes six times longer for inference. Our line-to-line searching strategy with pre-clustering can help parallelize the attention in IEA and CEA, leading to improved efficiency. We report the inference time of different resolutions in~\cref{fig5}. As the resolution scales up, the advantages of the line-to-line implementation will become more obvious.

\begin{table}
\centering
\renewcommand\arraystretch {1}
\resizebox{0.98\linewidth}{!}{
\begin{tabular}{@{}l|cccc@{}}
\toprule
Settings & Overall  $\downarrow$ & Depth Error  $\downarrow$  & MACs(G)  $\downarrow$ & Time(s) $\downarrow$ \\
\midrule
Point-to-Line & \textbf{0.290} & 5.774  & 25.66 & 3.06\\
Line-to-Line & 0.291 & \textbf{5.754}  & \textbf{0.586} & \textbf{0.46}\\
\bottomrule
\end{tabular}}
\vspace{2pt}
\caption{\textbf{Comparison of line-to-line and point-to-line algorithms.} ``MAC'' refers to the multiply–accumulate operations for ET.}
\label{Efficiency of line-to-line}
\vspace{-2pt}
\end{table}

\begin{table}
\centering
\setlength{\tabcolsep}{3pt}
\renewcommand\arraystretch {1} 
\resizebox{0.85\linewidth}{!}{
\begin{tabular}{@{}l|ccc@{}}
\toprule
Method & Overall(mm) $\downarrow$ & Depth Error(mm)  $\downarrow$& Param(M) $\downarrow$\\
\midrule
ET & \textbf{0.291} & \textbf{5.754} &  \textbf{1.09} \\
FMT & 0.303 & 6.064& 1.42 \\
DCN & 0.309 & 6.161 & 1.20 \\
ASPP & 0.320 & 7.227 & 1.26 \\                         
\bottomrule

\end{tabular}}
\vspace{4pt}
\caption{\textbf{Comparison of different feature enhancement methods.}}
\label{Different feature enhancement methods}
\vspace{-10pt}
\end{table}

\begin{figure}
    \includegraphics[scale=0.34]{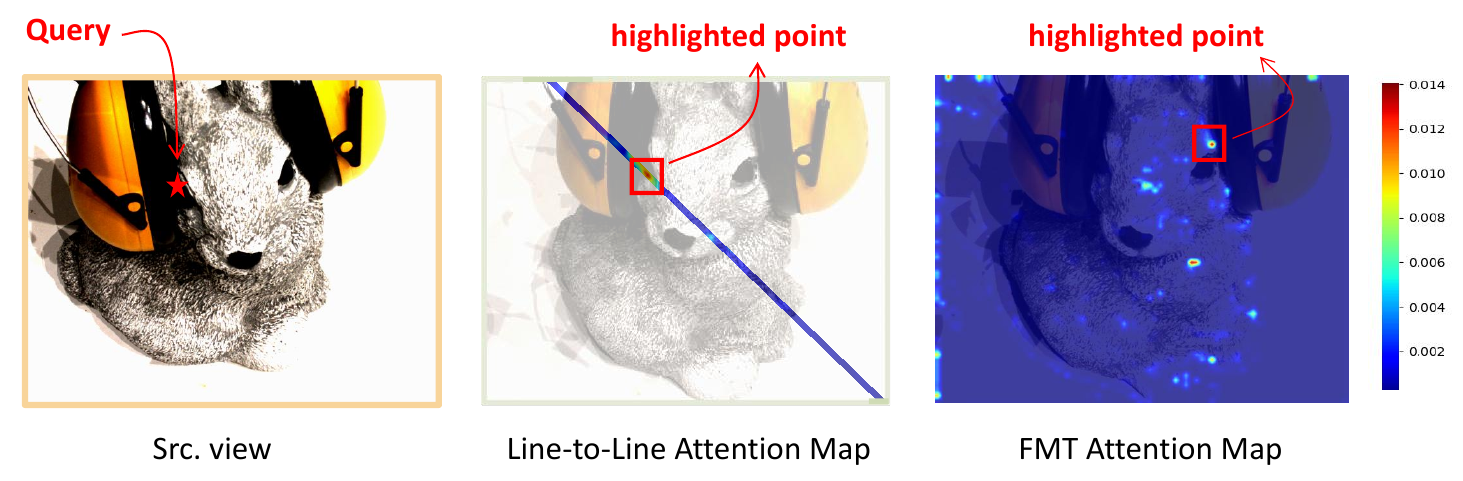}
    \caption{
    \textbf{Visualization of the attention map.} The attention map of our line-to-line Epipolar Transformer is between two corresponding epipolar lines and that of Feature Matching Transformer (FMT)~\cite{ding2022transmvsnet} is between a pixel and the other image. 
    Given a query point on an epipolar line of the source image, ET focuses on the region of its corresponding epipolar line while FMT considers the whole reference image. 
    The red box points to the area with the highest attention score for the attention maps of our line-to-line ET and FMT, respectively.
    }
    \label{fig6}
\end{figure}

\noindent \textbf{Different feature aggregation methods}
As shown in Table~\ref{Different feature enhancement methods}, we compare four feature aggregation methods: the proposed Epipolar Transformer (ET), Feature Matching Transformer(FMT)~\cite{ding2022transmvsnet}, Deformable Convolutional Networks (DCN)~\cite{dai2017deformable}, and Atrous Spatial Pyramid Pooling (ASPP)~\cite{chen2017rethinking}. Our Epipolar Transformer outperforms others in terms of both depth and construction metrics with the least parameters. This suggests that, based on the epipolar geometry as prior, the non-local feature aggregation on the epipolar line can be more effective and efficient for MVS, compared with other general-use feature aggregation paradigms. We further visualize the attention map in Fig.~\ref{fig6}. The feature matching transformer (FMT) in TransMVSNet~\cite{ding2022transmvsnet} generates attention on some irrelevant backgrounds and the points with similar textures, which can interfere with the representation. In contrast, by constraining the attention on the line, our ET can concentrate on the corresponding points.

\noindent \textbf{Results on other approaches}.
Our proposed Epipolar Transformer (ET) can serve as a plug-and-play module integrated into common learning-based MVS frameworks. Here, we apply ET to another representative baseline CasMVSNet~\cite{gu2020cascade}. Based on the original setting and implementation, ET is inserted into the FPN with the lowest resolution. As shown in Table~\ref{Plug-in}, CasMVSNet+ET improves the overall performance by $15\%$ compared with the original implementation on the DTU dataset. On the tanks and temples benchmark, CasMVSNet+ET achieves $8\%$ and $15\%$ relative improvements on the intermediate and advanced sets. The significant performance improvement across multiple datasets demonstrates the flexibility of our method.

\begin{table}
\centering
\renewcommand\arraystretch {1}
\resizebox{\linewidth}{!}{
\begin{tabular}{@{}l|ccc@{}}
\toprule
Method & DTU Overall  $\downarrow$ & Int. Mean$\uparrow$  & Adv. Mean $\uparrow$ \\
\midrule
CasMVSNet~\cite{gu2020cascade} & 0.355 & 56.84  & 31.12 \\
CasMVSNet+ET & \textbf{0.301} & \textbf{61.62}  & \textbf{35.65} \\
\bottomrule
\end{tabular}}
\vspace{2pt}
\caption{\textbf{Compatibility experiments on CasMVSNet.} ``+ ET'' indicates embedding our ET into the coarsest stage of its feature encoding structure.}
\label{Plug-in}
\vspace{-3pt}
\end{table}

\section{Conclusion}
In this paper, we introduce a non-local feature aggregation strategy based on epipolar constraints.
Specifically, we first propose an algorithm for searching epipolar pairs in two calibrated images. We then design an Intra-Epipolar Augmentation (IEA) module and a Cross-Epipolar Augmentation (CEA) module to mine non-local context within and across epipolar lines.
We pack these modules into a Transformer model termed Epipolar Transformer (ET) and integrate the ET into a baseline network to construct the ET-MVSNet. 
Evaluations and ablations verify the effectiveness of the proposed modules. In addition, we show that ET can also serve as a plug-and-play module integrated into other MVS methods.

\noindent \textbf{Acknowledgements.}
This work was funded in part by the National Natural Science Foundation of China (Grant No. U1913602) and supported by the Huawei Technologies CO., LTD.

{\small
\bibliographystyle{ieee_fullname}
\bibliography{main}
}

\clearpage
\appendix
\setcounter{equation}{0}
\renewcommand{\theequation}{a\arabic{equation}}

\newcounter{Sfigure}
\setcounter{Sfigure}{1}
\renewcommand{\thefigure}{A\arabic{Sfigure}}

\newcounter{STable}
\setcounter{STable}{1}
\renewcommand{\thetable}{A\arabic{STable}}

\section*{\centering\textbf{Supplementary Material}}
\section{More Details of Epipolar Pair Search}
\subsection{Mathematical Derivation}
In Sec.~\ref{Epipolar Pair Searching}, we obtain the expression of the corresponding epipolar lines (Eq.~\ref{equation 3}) from point homography transformation (Eq.~\ref{eq:equation 1}). Here, we will provide a detailed derivation of the process. To keep consistent, we illustrate the mathematical derivation with one source view.

Given a pixel $\mathbf{p}_r=(x_r,y_r,1)^{T}$ in the reference view, the corresponding pixel $\mathbf{p}_s$ in the source view is
\begin{equation}
\label{eq_homo}
\mathbf{p}_s(d) = \mathbf{K}_s [\mathbf{R}(\mathbf{K}_r^{-1}\mathbf{p}_rd)+\mathbf{t}]\,,
\end{equation}
where $d$ denotes the depth of the reference pixel; $\mathbf{R}$ and $\mathbf{t}$ indicate the rotation and translation matrices between the reference and the source view; $\mathbf{K}_r$ and $\mathbf{K}_s$ are the intrinsic matrices of the reference view and source view, respectively. For easy understand, \cref{eq_homo} can be expressed as
\begin{equation}
\label{eq1_convenience}
\mathbf{p}_s(d) =d_s*(x_s(d),y_s(d),1)^{T}= \mathbf{W} \mathbf{p}_rd + \mathbf{b}\,,
\end{equation}
where $\mathbf{W} = \mathbf{K}_s \mathbf{R} \mathbf{K}_r^{-1}$, $\mathbf{b} = \mathbf{K}_s\mathbf{t}$. We can further transform~\cref{eq1_convenience} into a coordinate form:
\begin{equation}
x_s(d)  = \frac{a_1 d + b_1}{a_3 d + b_3}\,,
y_s(d)  = \frac{a_2 d + b_2}{a_3 d + b_3}\,,
\label{eq2_copy}
\end{equation}
where $a_1 = w_{11} x_r + w_{12} y_r + w_{13}$, $a_2 = w_{21} x_r + w_{22} y_r + w_{23}$, $a_3 = w_{31} x_r + w_{32} y_r + w_{33}$;~$w_{ij}$ is an element of matrix $\mathbf{W}$, and $b_{i}$ is an element of vector $\mathbf{b}$. 

Since $\{a_i\}_{i=1}^3$ and $\{b_i\}_{i=1}^3$ are constants associated with the camera parameters and the coordinate of $\mathbf{p}_r$, the standard equation for the epipolar line $y_s(d)=kx_s(d)+b$ can be formulated as 
\begin{equation}
\label{equation3_copy}
\left\{
\begin{aligned}
k &= \frac{\Delta y_s(d)}{\Delta x_s(d)} = \frac{a_2 b_3 -a_3 b_2}{a_1 b_3 -a_3 b_1} \\
b &= y_s(0) - k x_s(0) = \frac{b_2}{b_3} - k \frac{b_1}{b_3}
\end{aligned}
\right.\,.
\end{equation}
Specifically, when ${\Delta x_s(d)}\rightarrow 0$, the stand equation for the epipolar line is $x_d(s) = k^{'}y_s(d) + b^{'}$, which can be formulated as

\begin{equation}
\label{equation3_extend}
\left\{
\begin{aligned}
k^{'} &= \frac{\Delta x_s(d)}{\Delta y_s(d)} = \frac{a_1 b_3 -a_3 b_1}{a_2 b_3 -a_3 b_2} \\
b^{'} &= x_s(0) - k^{'} y_s(0) = \frac{b_1}{b_3} - k^{'} \frac{b_2}{b_3}
\end{aligned}
\right.\,.
\end{equation}

\subsection{Discussion about Quantification}
we quantify the pre-calculated $k$ and $b$ by rounding as
\begin{equation}
\label{quatification}
\left\{
\begin{aligned}
k &= s_k*\mathrm{round}(\frac{k}{s_k}) \\
b &= s_b*\mathrm{round}(\frac{b}{s_b})
\end{aligned}
\right.\,,
\end{equation}
where $s_k$ and $s_b$ are the hyperparameters for rounding, and quantization precision~($k,b$) depends on the precision of them. To explore the effect of quantization precision for epipolar pair search, we list common precision combinations of $s_k$ and $s_b$ in~\cref{quantification accuracy}.



The quantization precision of both $k$ and $b$ will affect the effect of epipolar pair search, and thus affect the performance. Besides, the quantization precision also influences the efficiency of epipolar pair search, as finer quantization precision leads to more clusters and vice versa. Considering the effectiveness and efficiency, we choose the precision combination of $s_k=0.1$ and $s_b=10$ in our implementation.


\begin{table}[t]
\centering
\renewcommand\arraystretch {1}
\resizebox{0.98\linewidth}{!}{
\begin{tabular}{@{}cc|ccccc@{}}
\toprule
$s_k$ & $s_b$  & ACC.(mm) $\downarrow$ & Comp.(mm) $\downarrow$ & Overall(mm) $\downarrow$ \\
\midrule
1 & 0.1 &  0.327 & 0.267 & 0.297  \\
1 & 1 & 0.328 & 0.260 & 0.294  \\
1 & 10 &  0.325 & 0.263 & 0.294  \\
0.1 & 0.1 &\textbf{0.324} & 0.263 & 0.294 \\
0.1 & 1 & 0.325 & 0.257 & \textbf{0.291} \\
0.1 & 10 & 0.329 & \textbf{0.253} & \textbf{0.291}  \\
0.01 & 0.1 &  0.325 & 0.271 & 0.298 \\
0.01 & 1 & 0.332 & 0.260 & 0.296 \\
0.01 & 10 & 0.327 & 0.265 & 0.296 \\

\bottomrule
\end{tabular}}
\vspace{2pt}
\caption{\textbf{Comparison of different quantification precision.}}
\label{quantification accuracy}
\addtocounter{STable}{1}
\end{table}

\section{Efficiency Comparison}
We empirically study the efficiency of the point-to-line and the line-to-line implementations in our ablation studies (Sec.~\ref{Ablation Study}). Here, we analyze their complexity theoretically. 
In addition, we compare the global aggregation strategy: plane-to-plane~(Linear), which is applied in TransMVSNet~\cite{ding2022transmvsnet}.
\subsection{Theoretical Efficiency Comparison}

Given $Q \in R^{B \times N_1 \times C}$, $K \in R^{B \times N_2 \times C}$, and $V \in R^{B \times N_2 \times C}$, the computational complexity of the vanilla Transformer~\cite{vaswani2017attention} is $B(9N_1C^2+2N_2C^2+2N_1N_2C)$. The computational complexity of the linear Transformer~\cite{katharopoulos2020transformers} is $B(10N_1C^2+3N_2C^2)$. 

For line-to-line and point-to-line, the computational complexity depends on the number of epipolar lines as well.
Specifically, suppose there are $M$ corresponding epipolar lines and the average number of pixels on an epipolar line is $S$. For point-to-line, $B = HW$, $N_1 = 1$, $N_2 = S$, its computational complexity is $HW(9C^2+2SC^2+2SC)$.
For line-to-line, $B = M$, $N_1 = S$, $N_2 = S$, its computational complexity is $M(11SC^2+2S^2C)$.
It is worth noting that $M$ and $S$ are of the same order of magnitude as $H$ and $W$. They are usually smaller because the epipolar lines only exist in the common view of the two images.
For the plane-to-plane in the form of linear Transformer implementation, $B=1$, $N_1 = N_2 = HW$, its computational complexity is $13HWC^2$.

For a more intuitive comparison, we set $H=80$, $W=64$, $C=64$, $S=30$, $M=30$: the computational complexity of point-to-line is 1.5G;
the computational complexity of line-to-line is 0.04G; and the computational complexity of plane-to-plane (linear) is 0.27G. 

\subsection{Empirical Efficiency Comparison}
We report the inference time to compare different aggregate ways in practice.
As shown in Table~\ref{Different ways of information aggregation}, ``line-to-line'' is a more efficient and effective way to aggregate information, which maintains the highest performance while being the lowest in terms of time and memory consumption.

\begin{table}[t]
\centering
\renewcommand\arraystretch {1}
\resizebox{0.98\linewidth}{!}{
\begin{tabular}{@{}l|ccc@{}}
\toprule
mothod & Overall(mm)  $\downarrow$ & Time(ms) $\downarrow$ & Memory(MB) $\downarrow$\\
\midrule
Line-to-line & 0.291 & \textbf{2.0} & \textbf{2769}  \\
Point-to-line & \textbf{0.290} &3.5 & 4207  \\
Plane-to-plane (Linear) & 0.303 & 3.9 & 2997  \\
\bottomrule
\end{tabular}}
\vspace{2pt}
\caption{\textbf{Comparison of different ways of information aggregation.} ``Line-to-line'' refers to the information aggregation between epipolar pairs. ``Point-to-line'' refers to a point interacting with its corresponding epipolar line. ``Plane-to-plane (Linear) refers to the information aggregation between two whole images in the form of linear Transformer~\cite{katharopoulos2020transformers} implementation. 
``Time'' refers to inference time through one Transformer~(for $864 \times 1152$ images).}
\label{Different ways of information aggregation}
\addtocounter{STable}{1}
\end{table}

\section{Additional Ablation Studies}
\subsection{Number of Blocks}
Table~\ref{number of layers} shows the impact of different block numbers of the Intra-Epipolar Augmentation (IEA) and Cross-Epipolar Augmentation (CEA).
As the number increases, no performance gain is obtained, which suggests that one block is sufficient for non-local feature aggregation.
Since more blocks lead to larger computation overhead, $N_a$ is set to 1 in our implementation.

\begin{table}[t]
\centering
\renewcommand\arraystretch {1}
\resizebox{0.98\linewidth}{!}{
\begin{tabular}{@{}l|cccccc@{}}
\toprule
$N_a$ & ACC.(mm) $\downarrow$ & Comp.(mm) $\downarrow$ & Overall(mm) $\downarrow$ & Time(s) $\downarrow$ & Param(M) $\downarrow$ \\
\midrule
1 & 0.329 & \textbf{0.253} & 0.2910 & \textbf{0.46} & \textbf{1.09} \\
2 & \textbf{0.327} & 0.254 & \textbf{0.2905} & 0.47 & 1.16 \\
3 & \textbf{0.327} & 0.254 
 & \textbf{0.2905} & 0.47 & 1.23 \\
\bottomrule
\end{tabular}}
\vspace{2pt}
\caption{\textbf{Ablation study on the number of IEA and CEA blocks.}}
\label{number of layers}
\addtocounter{STable}{1}
\end{table}

\subsection{Spatial Positional Encoding}
In Epipolar Transformer (ET), we apply Positional Encoding (PE) to add spatial positional information to feature sequences. We compare different positional encoding implementations in Table~\ref{PE}. Positional encoding is necessary and the performance of ``learnable'' and ``sine'' positional encoding are similar.
 Since ``sine'' positional encoding is parameter-free, we used ``sine'' positional encoding in our implementation.

\begin{table}[t]
\centering
\renewcommand\arraystretch {1}
\resizebox{0.98\linewidth}{!}{
\begin{tabular}{@{}l|cccc@{}}
\toprule
PE & ACC.(mm) $\downarrow$ & Comp.(mm) $\downarrow$ & Overall(mm) $\downarrow$ \\
\midrule
w/o & \textbf{0.329} & 0.260 & 0.295 \\
learnable & 0.330 &0.254 & 0.292 \\
sine & \textbf{0.329} & \textbf{0.253} & \textbf{0.291}  \\

\bottomrule
\end{tabular}}
\vspace{2pt}
\caption{\textbf{Ablation study on spatial positional encoding.}}
\label{PE}
\vspace{1pt}
\addtocounter{STable}{1}
\end{table}

\begin{figure}
    \centering
    \includegraphics[width=0.46\textwidth,height=0.22\textheight]{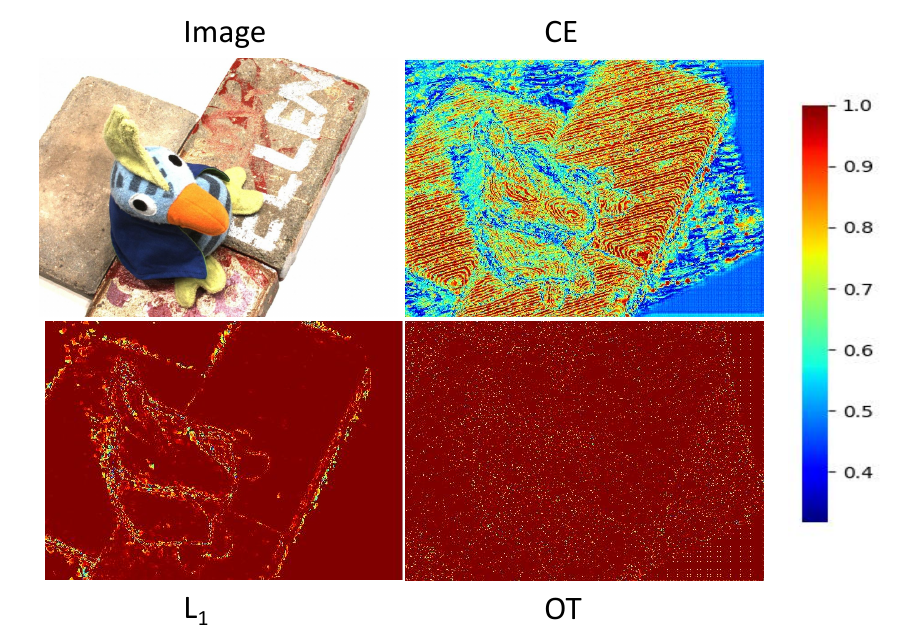}
    \caption{\textbf{Visualization of confidence maps generated through different loss functions.}}
    \label{conf_map}
    \vspace{1pt}
    \addtocounter{Sfigure}{1}
\end{figure}

\subsection{Loss Function}
In Sec.~\ref{Implementation Details}, we formulate depth estimation as a classification problem and apply cross-entropy loss as the loss function of ET-MVSNet. As shown in Table~\ref{loss}, we compare different loss functions. The experimental results indicate that the cross-entropy loss achieves the highest overall metric, which better balances accuracy and completeness. Besides, as shown in Fig.~\ref{conf_map}, the confidence map generated through ``CE'' is more advantageous for filtering outliers and obtaining more accurate point clouds.

\begin{table}[t]
\centering
\renewcommand\arraystretch {1}
\resizebox{0.98\linewidth}{!}{
\begin{tabular}{@{}l|cccc@{}}
\toprule
Loss & ACC.(mm) $\downarrow$ & Comp.(mm) $\downarrow$ & Overall(mm) $\downarrow$  \\
\midrule
CE & \textbf{0.329} & 0.253 & \textbf{0.291} \\
OT & 0.383 & \textbf{0.223} & 0.303 \\
$L_1$ & 0.367 & 0.245 & 0.306 \\

\bottomrule
\end{tabular}}
\vspace{2pt}
\caption{\textbf{Ablation study on different loss functions.}``CE'' refers to the commonly used cross-entropy loss and ``OT'' refers to the Wasserstein loss computed by optimal transport, where the depth estimation is regarded as a classification problem. ``$L_1$'' refers to the average absolute value error where the depth estimation is regarded as a regression problem.}
\label{loss}
\vspace{1pt}
\addtocounter{STable}{1}
\end{table}

\begin{figure}[t]
    \includegraphics
    [width=0.45\textwidth]
    {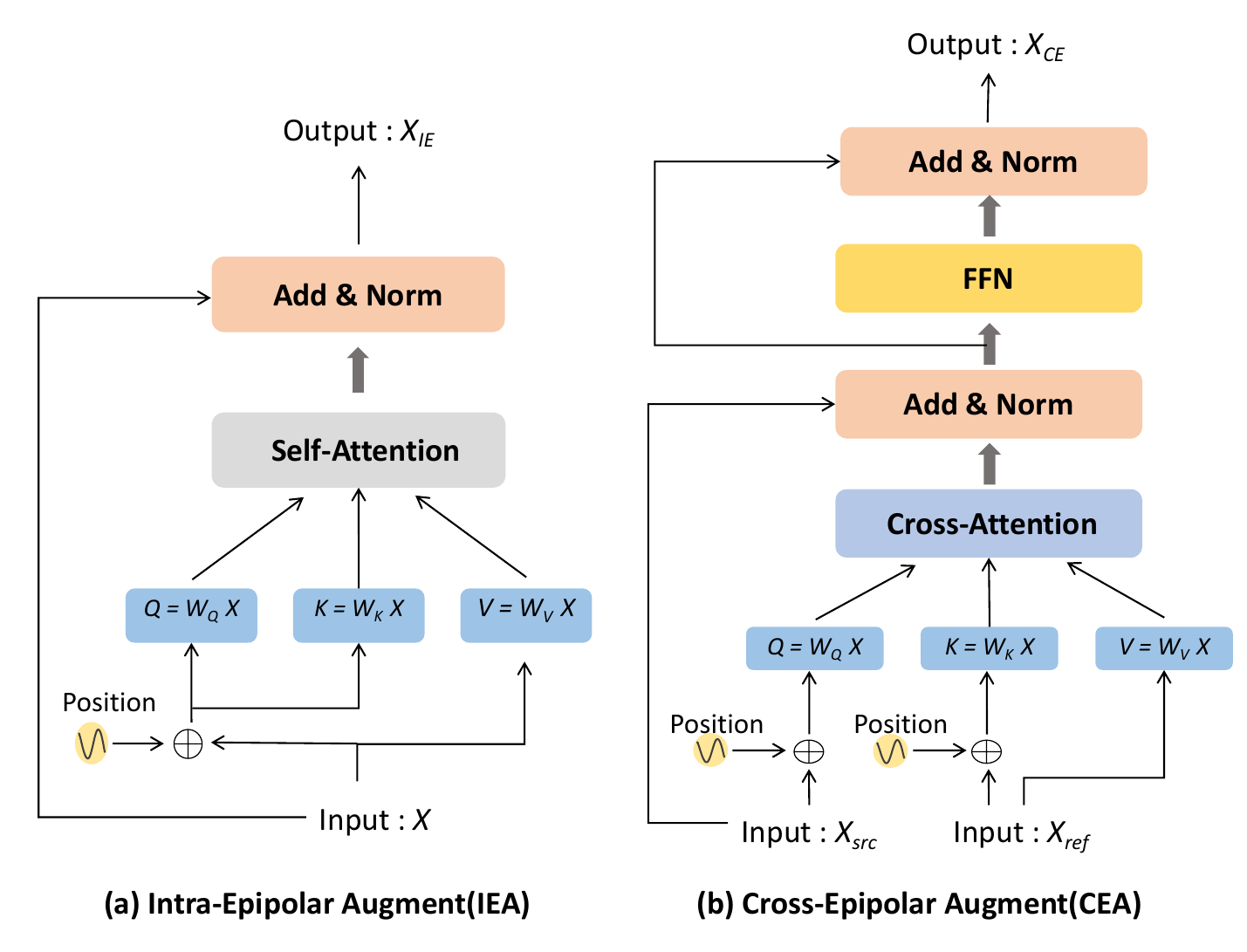}
    \vspace{2pt}
    \caption{\textbf{The structure of IEA and CEA.} IEA is based on the self-attention mechanism while CEA is based on the cross-attention mechanism.} 
    \label{IEAandCEA}
    \addtocounter{Sfigure}{1}
\end{figure}

\begin{table}[t]
\centering
\renewcommand\arraystretch {1}
\resizebox{0.98\linewidth}{!}{
\begin{tabular}{@{}l|cccc@{}}
\toprule
Order & ACC.(mm) $\downarrow$ & Comp.(mm) $\downarrow$ & Overall(mm) $\downarrow$  \\
\midrule
IEA+CEA & 0.329 & \textbf{0.253} & \textbf{0.291} \\
CEA+IEA & \textbf{0.327} & 0.257 & 0.292 \\

\bottomrule
\end{tabular}}
\vspace{2pt}
\caption{\textbf{Ablation study on different orders of IEA and CEA.}}
\label{order}
\addtocounter{STable}{1}
\end{table}

\subsection{Order of IEA and CEA}
As shown in Fig.~\ref{IEAandCEA}, the Intra-Epipolar Augmentation (IEA) and Cross-Epipolar Augmentation (CEA) modules perform information aggregation within and across epipolar lines, respectively. We explore the order of IEA and CEA in Table~\ref{order}. The order of IEA and CEA has little impact on the final performance. In our implementation, IEA is executed first followed by CEA.

\section{Depth Map Fusion}
As described in the main text, the predicted depth maps of multiple views are filtered and fused into a point cloud. 
Previous MVS methods always choose the suitable fusion method. In the paper, we follow the commonly used dynamic checking strategy~\cite{yan2020dense} for depth filtering and fusion on both DTU dataset~\cite{dtu} and Tanks and Temples benchmark~\cite{tanks}.

On the DTU dataset, we filter the confidence map of the last stage with a confidence threshold($0.55$) to measure photometric consistency. For geometry consistency, we use a strict standard, as shown below.
\begin{equation}
\label{dypcd}
err_c < thresh_c\,,
err_d < log(thresh_d)\,,
\end{equation}
where $err_c$ and $err_d$ denote the reprojection coordinate error and relative error of reprojection depth, respectively. $thresh_c$ and $thresh_d$ denote thresholds for $err_c$ and $err_d$, respectively. In addition, we adopt the normal(pcd) fusion method~\cite{schonberger2016pixelwise}, and our method can achieve $0.298$ on the ``overall'' metric. 

On the Tanks and Temples benchmark, We follow~\cite{unimvs} to adjust hyperparameters for each scene including confidence thresholds, geometric thresholds, etc. For benchmarking on the advanced set of Tanks and Temples , the number of depth hypotheses in the coarsest stage is changed from 8 to 16. And we use the model trained on the DTU dataset to reconstruct the ``Horse'' scene, and then use the fine-tuned model on BlendedMVS dataset~\cite{yao2020blendedmvs} to reconstruct other scenes.

\section{More Visualization Results}
\begin{figure*}[t]
    \centering
    \includegraphics[width=0.95\textwidth,height=0.6\textheight]{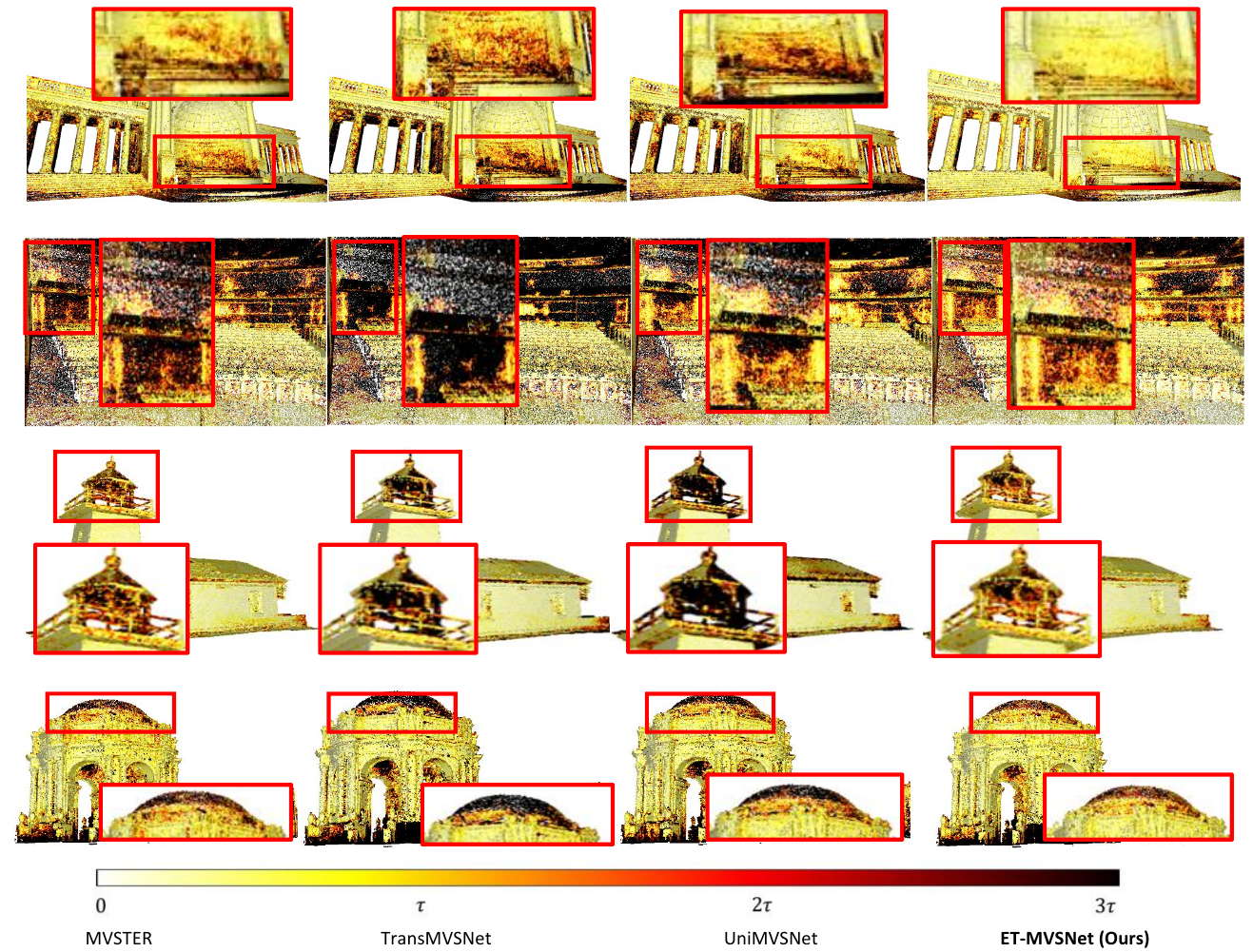}
    \caption{\textbf{Visualization comparison with state-of-the-art methods~\cite{wang2022mvster,ding2022transmvsnet,unimvs} on Tanks and Temples benchmark.} From top to bottom is the Recall of the scene of the Temple($\tau = 15mm$), Auditorium ($\tau = 10mm$), Lighthouse ($\tau = 5mm$), and Palace ($\tau = 30mm$), respectively.} 
    \label{TT}
    \addtocounter{Sfigure}{1}
\end{figure*}

\noindent \textbf{Qualitative Analysis}.
The Qualitative results of Tanks and Temples benchmark~\cite{tanks} are shown in Fig.~\ref{TT}. Compared with other state-of-the-art methods~\cite{wang2022mvster,ding2022transmvsnet,unimvs}, our method can reconstruct more details in some challenging areas, such as surfaces with weak and repetitive textures.

\begin{figure*}[t]
    \includegraphics[width=0.97\textwidth,height=0.22\textheight]{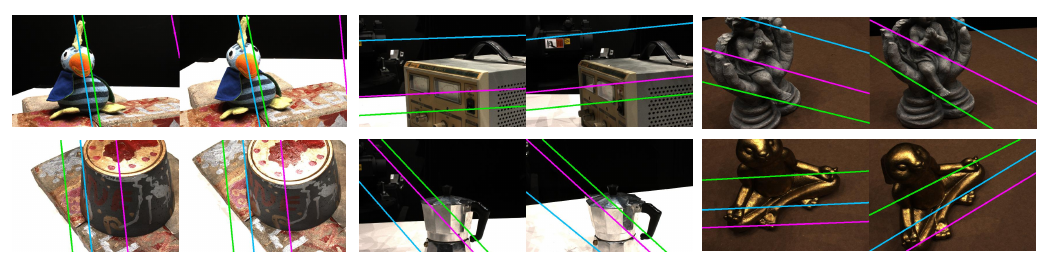}
    \caption{\textbf{Visualization of epipolar line pairs.} The same color indicates the corresponding epipolar line pair.} 
    \label{epipolarpairs}
    \addtocounter{Sfigure}{1}
\end{figure*}

\noindent \textbf{Epipolar Pairs}. We visualize some epipolar line pairs searched by our algorithm in Fig.~\ref{epipolarpairs}. 
In the case of large differences, pixels with the same semantic information are still on the same epipolar line pair, indicating the effectiveness of our algorithm.

\begin{figure*}
    \centering
    
    \includegraphics[width=0.90\textwidth,height=0.8\textheight]{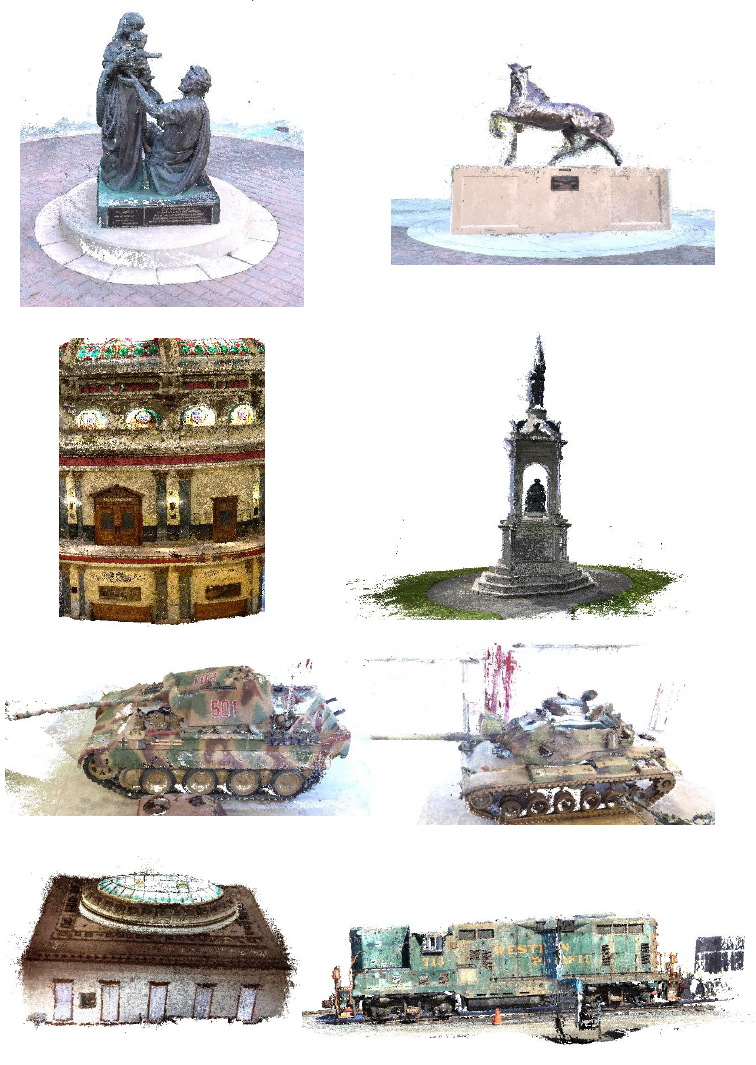}
    \caption{\textbf{Point clouds reconstructed by ET-MVSNet on Tanks and Temples benchmark~\cite{tanks}.}}
    \label{TT_cloudpoints}
    \addtocounter{Sfigure}{1}
\end{figure*}

\noindent \textbf{More Point Cloud Results}. 
More visualization results of our model are shown in Fig.~\ref{TT_cloudpoints}, which contains point clouds of Tanks and Temples benchmark~\cite{tanks}.

\end{document}